\documentclass[11pt]{article}
\usepackage[preprint]{acl}

\usepackage{times}
\usepackage{latexsym}
\usepackage[T1]{fontenc}
\usepackage{microtype}
\usepackage{inconsolata}
\usepackage{float}
\usepackage{graphicx}
\usepackage{booktabs}
\usepackage{amsmath}
\usepackage{amssymb}
\usepackage{algorithm}
\usepackage{algpseudocode}
\usepackage{multirow}
\usepackage{subcaption}
\usepackage{tikz}
\usetikzlibrary{calc,arrows.meta,positioning}
\usepackage{url}
\usepackage{placeins}

\title{NAVIRA: Decoupled Stochastic Remasking for Masked Diffusion Language Models}

\author{
Andrey Fomenko$^{1}$,
Maksim Kryzhanovskiy$^{1,2}$,
Svetlana Glazyrina$^{1}$,
Roman Ischenko$^{1,2}$ \\
\\
$^{1}$ Lomonosov Moscow State University, Moscow, Russia \\
$^{2}$ Institute for Artificial Intelligence, Lomonosov Moscow State University, Moscow, Russia
}

\begin{document}
\maketitle

\begin{abstract}
Masked diffusion language models generate text by iteratively unmasking many tokens in parallel, but this speed comes with a correction problem: tokens generated in the same step are predicted from marginal distributions, and early local dependency errors can later contaminate the context. PRISM addresses this by learning token-level quality scores and remasking unreliable tokens, but its inference rule is coupled: the same forward pass both detects low-quality tokens and computes logits for their replacements, so the erroneous tokens still condition regeneration. We propose NAVIRA, an inference-time decoding policy that separates these two operations and samples remasking positions stochastically. A first forward pass scores tokens; selected tokens are masked; a second forward pass regenerates from the cleaned context. Temperature-controlled remasking reduces repeated correction of the same positions and balances fluency against diversity. In controlled experiments with a 170M masked diffusion language model, decoupling improves fluency, while scheduled stochastic remasking preserves entropy and achieves stronger LLM-judge scores under larger forward-pass budgets. These results show that remasking policy, not only the learned quality signal, is central to reliable masked-diffusion text generation.
\end{abstract}

\begin{figure*}[h]
\centering
\resizebox{0.96\textwidth}{!}{%
\begin{tikzpicture}[
    >=Latex,
    every node/.style={align=center},
    stage/.style={
        draw,
        rounded corners=3pt,
        fill=gray!4,
        line width=0.45pt,
        minimum width=4.20cm,
        minimum height=1.90cm
    },
    title/.style={font=\scriptsize\bfseries},
    note/.style={font=\scriptsize, text=gray!70!black},
    token/.style={
        draw,
        rounded corners=1.5pt,
        minimum width=0.62cm,
        minimum height=0.34cm,
        inner sep=1.2pt,
        font=\scriptsize,
        line width=0.35pt
    },
    kept/.style={token, fill=blue!12, draw=blue!55!black},
    wrong/.style={token, fill=orange!25, draw=red!70!black, line width=0.6pt},
    mask/.style={token, fill=red!10, draw=red!65!black},
    fresh/.style={token, fill=green!14, draw=green!55!black},
    scoregood/.style={
        draw=green!55!black,
        fill=green!10,
        rounded corners=1pt,
        minimum width=0.52cm,
        minimum height=0.30cm,
        inner sep=1.2pt,
        font=\scriptsize
    },
    scorebad/.style={scoregood, draw=red!65!black, fill=red!10},
    arrow/.style={->, line width=0.70pt},
    bluearrow/.style={arrow, blue!70!black},
    redarrow/.style={arrow, red!70!black},
    lab/.style={font=\scriptsize}
]

\coordinate (X)  at (0,0);
\coordinate (Q)  at (5.10,0);
\coordinate (P)  at (10.20,0);
\coordinate (Y)  at (0,-3.00);
\coordinate (F)  at (5.10,-3.00);
\coordinate (C)  at (10.20,-3.00);

\foreach \c in {X,Q,P,C,F,Y} {
    \node[stage] at (\c) {};
}

\node[title] at ($(X)+(0,0.56)$) {current state};
\node[kept]  at ($(X)+(-1.14,0.07)$) {I};
\node[kept]  at ($(X)+(-0.38,0.07)$) {saw};
\node[wrong] at ($(X)+(0.38,0.07)$) {a};
\node[kept]  at ($(X)+(1.14,0.07)$) {apple};
\node[lab] at ($(X)+(0,-0.46)$) {$x^{(t)}$};
\node[note, text=red!70!black] at ($(X)+(0,-0.74)$) {wrong article contaminates context};

\node[title] at ($(Q)+(0,0.56)$) {first forward: scoring};
\node[scoregood] at ($(Q)+(-1.14,0.07)$) {.94};
\node[scoregood] at ($(Q)+(-0.38,0.07)$) {.89};
\node[scorebad]  at ($(Q)+(0.38,0.07)$) {.16};
\node[scoregood] at ($(Q)+(1.14,0.07)$) {.91};
\node[lab, text=teal!70!black] at ($(Q)+(0,-0.46)$) {$s_i=g_\phi(x^{(t)})_i$};
\node[note] at ($(Q)+(0,-0.74)$) {lower score $\Rightarrow$ stronger remask candidate};

\node[title] at ($(P)+(0,0.56)$) {stochastic remasking};
\node[lab, text=red!70!black] at ($(P)+(0,0.17)$) {$\pi_i(\tau_t)\propto\exp(-s_i/\tau_t)$};
\node[lab] at ($(P)+(0,-0.22)$) {$T_t \sim \pi(\tau_t)$ over clean tokens};
\draw[->, line width=0.35pt] ($(P)+(-1.45,-0.68)$) -- ($(P)+(-1.45,-0.36)$);
\draw[->, line width=0.35pt] ($(P)+(-1.45,-0.68)$) -- ($(P)+(-0.45,-0.68)$);
\draw[blue!70!black, line width=0.65pt]
    ($(P)+(-1.34,-0.64)$)
    .. controls ($(P)+(-1.08,-0.64)$) and ($(P)+(-0.92,-0.61)$) ..
    ($(P)+(-0.78,-0.51)$)
    .. controls ($(P)+(-0.66,-0.43)$) and ($(P)+(-0.56,-0.38)$) ..
    ($(P)+(-0.49,-0.37)$);
\node[lab, text=blue!70!black, anchor=west] at ($(P)+(-0.25,-0.56)$) {scheduler sets $\tau_t$};

\node[title] at ($(C)+(0,0.56)$) {remask state};
\node[kept] at ($(C)+(-1.14,0.07)$) {I};
\node[kept] at ($(C)+(-0.38,0.07)$) {saw};
\node[mask] at ($(C)+(0.38,0.07)$) {[M]};
\node[kept] at ($(C)+(1.14,0.07)$) {apple};
\node[lab] at ($(C)+(0,-0.46)$) {$\tilde{x}^{(t)}$};
\node[note, text=red!70!black] at ($(C)+(0,-0.74)$) {sampled $T_t$ is masked before generation};

\node[title, text=blue!70!black] at ($(F)+(0,0.56)$) {second forward: generation};
\node[lab] at ($(F)+(0,0.07)$) {$f_\theta(\cdot\mid\tilde{x}^{(t)})$};
\node[note] at ($(F)+(0,-0.42)$) {separate unmasking pass};
\node[note, text=blue!70!black] at ($(F)+(0,-0.74)$) {detection and regeneration are decoupled};

\node[title] at ($(Y)+(0,0.56)$) {refined state};
\node[kept]  at ($(Y)+(-1.14,0.07)$) {I};
\node[kept]  at ($(Y)+(-0.38,0.07)$) {saw};
\node[fresh] at ($(Y)+(0.38,0.07)$) {an};
\node[kept]  at ($(Y)+(1.14,0.07)$) {apple};
\node[lab] at ($(Y)+(0,-0.46)$) {$x^{(t+1)}$};
\node[note, text=green!50!black] at ($(Y)+(0,-0.74)$) {second forward recovers ``an apple''};

\draw[arrow] ($(X)+(2.12,0.07)$) -- ($(Q)+(-2.12,0.07)$);
\draw[arrow] ($(Q)+(2.12,0.07)$) -- ($(P)+(-2.12,0.07)$);
\draw[redarrow] ($(P)+(0,-0.95)$) -- ($(C)+(0,0.95)$);
\draw[bluearrow] ($(C)+(-2.12,0.07)$) -- ($(F)+(2.12,0.07)$);
\draw[bluearrow] ($(F)+(-2.12,0.07)$) -- ($(Y)+(2.12,0.07)$);

\node[lab] at (5.10,-4.42) {
    \textcolor{blue!65!black}{\rule{0.55em}{0.55em}} kept context
    \quad
    \textcolor{orange!75!black}{\rule{0.55em}{0.55em}} low-quality token
    \quad
    \textcolor{red!65!black}{\rule{0.55em}{0.55em}} mask / remask
    \quad
    \textcolor{green!55!black}{\rule{0.55em}{0.55em}} generated token
};

\end{tikzpicture}%
}
\caption{Overview of the proposed NAVIRA step. A first forward pass computes quality scores on the current state, a temperature-controlled scheduler samples positions to remask, and a separate second forward pass recomputes unmasking logits from the remasked context.}
\label{fig:method-overview}
\end{figure*}
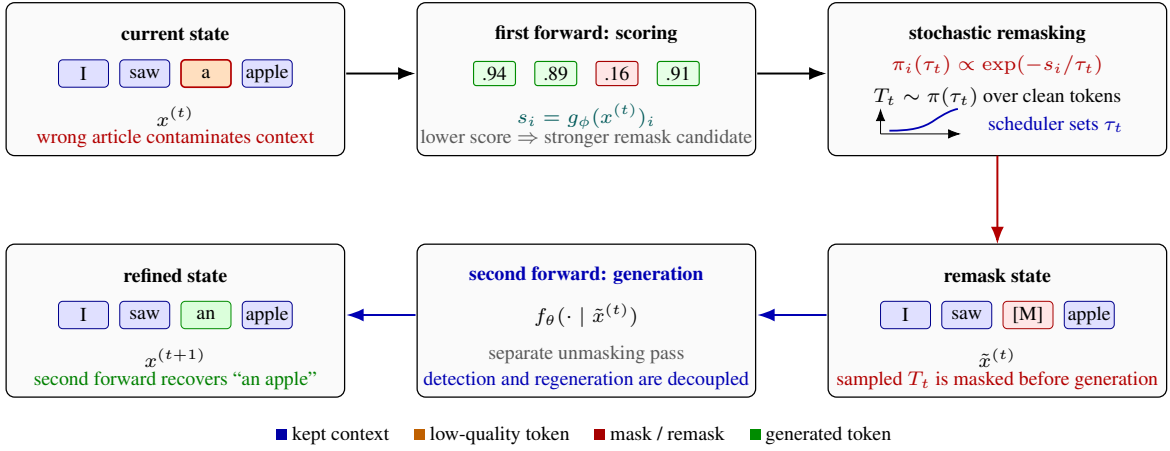

\section{Introduction}

Masked diffusion models (MDMs) offer an alternative to autoregressive text generation by iteratively denoising a partially masked sequence and updating multiple positions in parallel. This non-left-to-right generation process enables bidirectional conditioning and can substantially reduce decoding latency. However, parallel token updates also introduce a characteristic failure mode: tokens generated in the same step are predicted from marginal distributions and may fail to respect local dependencies. Once such errors enter the partially generated sequence, they can influence subsequent predictions and degrade later refinement.

Inference-time remasking provides a natural mechanism for addressing this issue. In particular, PRISM augments a pretrained MDM with a token-level quality head and uses the resulting scores to identify unreliable tokens for correction \citep{prism}. This makes it possible to revisit previously generated positions rather than permanently committing to early decisions. Despite this advantage, the standard PRISM decoding rule remains coupled: the same forward pass is used both to score tokens and to produce generation logits. Consequently, tokens selected as low-quality still participate in the context used to generate their replacements, allowing erroneous tokens to contaminate the very predictions intended to correct them.

We propose NAVIRA, a simple inference-time modification of PRISM based on two design choices. First, we decouple remasking from generation. A first forward pass estimates token quality and selects unreliable positions; these positions are then masked out, and a second forward pass recomputes generation logits from the corrected context. This ensures that tokens marked for correction no longer condition the regeneration step. Second, we replace deterministic top-$K$ remasking with stochastic remasking: positions are sampled from a temperature-controlled distribution derived from token-quality scores. This reduces repeated selection of the same low-scoring positions and provides a controllable mechanism for balancing correction and exploration.

Our study isolates the effect of these inference-time choices while keeping the pretrained backbone and PRISM training objective fixed. We compare deterministic and stochastic remasking variants, fixed and scheduled temperatures, and non-learned uncertainty-based baselines. Across unconditional text generation experiments, decoupled remasking improves fluency by using a cleaner context for regeneration, while stochastic remasking helps preserve diversity and mitigates entropy collapse under larger decoding budgets. LLM-based evaluation further suggests that scheduled stochastic remasking improves perceived generation quality, highlighting the importance of decoding policy design for reliable masked-diffusion text generation.

Our work presents the following contributions to the field of Diffusion Language Models (DLMs) and inference approaches:

\begin{itemize}
    \item We identify a coupling issue in PRISM-style inference and propose a two-pass decoding policy that separates token-quality scoring from regeneration, allowing correction to condition on a cleaned context.

    \item We introduce temperature-controlled stochastic remasking for MDM inference, including fixed-temperature and scheduled variants that trade off focused correction against diversity-preserving exploration.

    \item We provide controlled ablations under matched forward-pass budgets showing that decoupling improves fluency, while scheduled stochastic remasking mitigates entropy collapse and improves LLM-based generation quality.
\end{itemize}

\section{Related Work}

\paragraph{Masked diffusion language models.}
Discrete diffusion models, and masked diffusion models in particular, have gained significant interest as an approach to text generation \citep{austin2021d3pm,sedd,mdlm,llada,shi2024simplifiedgeneralizedmaskeddiffusion}. Unlike autoregressive models, they generate tokens in parallel and refine a partially masked sequence over multiple steps. However, errors made at early stages of generation can propagate across iterations and negatively affect later predictions. This line is also connected to earlier iterative masked generation, such as Mask-Predict \citep{ghazvininejad2019maskpredictparalleldecodingconditional}, and to recent analyses that view absorbing-state MDMs as modeling clean-token conditional distributions or time-agnostic masked models \citep{ou2024absorbingdiscretediffusionsecretly,zheng2024maskeddiffusionmodelssecretly}.

\paragraph{Self-correction and remasking.}
A growing line of work explores how to revise low-quality tokens during inference. In diffusion-based decoders such as Dream 7B and LLaDA \citep{dream7b,llada}, remasking is typically restricted to tokens that were updated in the current iteration, while tokens generated in earlier steps remain fixed. This design simplifies the decoding process but limits the ability to revisit and correct earlier mistakes.

In contrast, approaches such as ReMDM \citep{remdm} allow remasking over the full set of revealed tokens, enabling the model to revise decisions made in previous iterations. These methods typically rely on heuristic rules or confidence proxies derived from model outputs. While simple and effective, such proxies are not explicitly trained to reflect token correctness and may become unreliable as the sequence evolves. Related inference-time corrector methods include forward--backward correctors for discrete denoising models \citep{campbell2022continuoustimeframeworkdiscrete} and discrete flow matching \citep{gat2024discreteflowmatching}.

PRISM addresses this limitation by learning token-level quality scores jointly with the backbone, enabling the model to estimate both token distributions and correction signals within a unified framework \citep{prism}. This design also simplifies fine-tuning, as it allows adapting the model to a specific task by training a lightweight auxiliary head without full fine-tuning of the backbone. Related quality- or correction-based methods include Token-Critic \citep{lezama2022improvedmaskedimagegeneration}, RemeDi \citep{huang2025dontsettleearlyselfreflective}, Informed Correctors \citep{zhao2024informedcorrectorsdiscrete}, and GIDD \citep{vonrutte2025generalizedinterpolatingdiscretediffusion}; our focus is complementary, as we keep the backbone and PRISM objective fixed and study the inference-time remasking policy.

\paragraph{Decoding strategies in diffusion language models.}
A key design choice in masked diffusion language models is how positions are selected and updated during iterative decoding. Existing approaches differ in whether they restrict updates to recently generated tokens or allow revisiting the entire sequence. Many diffusion decoders adopt constrained update rules, where only tokens modified in the current step are eligible for further refinement, which simplifies the decoding process but limits the ability to correct earlier mistakes.

More flexible formulations allow remasking over all revealed tokens, enabling global refinement across iterations, but typically rely on heuristic selection rules or simple confidence signals. This highlights the importance of token selection policies as a core component of diffusion-based generation. Recent work further studies which masked positions should be revealed and how they should be grouped, including path planning for masked diffusion sampling \citep{peng2025pathplanningmaskeddiffusion} and dilated scheduling for faster MDM decoding \citep{luxembourg2025planspeeddilatedscheduling}. Our stochastic remasking is also motivated by the broader observation that overly deterministic text decoding can reduce diversity and lead to degeneration \citep{holtzman2019curiouscaseneuraltext}.

\paragraph{LLM-based generation evaluation.}
Our evaluation is also related to LLM-as-a-judge protocols for open-ended generation quality, such as MT-Bench and Chatbot Arena \citep{zheng2023judgingllmasajudgemtbenchchatbot}. We use these judgments as a complementary diagnostic to perplexity and entropy rather than as the sole evaluation criterion.

\section{Preliminaries}

\subsection{Masked Diffusion Models}

Let $x = (x_1,\dots,x_L) \in V^L$ denote a clean discrete sequence of length $L$ over vocabulary $V$, and let $m$ denote the mask token. A masked diffusion model is trained to recover clean tokens from partially masked versions of the sequence. Given a corrupted sequence $z \in (V \cup \{m\})^L$, obtained by masking a subset of positions in $x$, the model $f_\theta$ predicts, for each position $i$, a distribution over possible tokens:

\begin{equation}
    f_\theta(\cdot \mid z)_i \approx p(x_i \mid z).
\end{equation}

At inference time, decoding starts from a fully masked sequence and proceeds iteratively. At each step, a subset of masked positions is selected and filled using the model predictions. The choice of which positions to reveal is flexible and plays a key role in generation quality.

\subsection{PRISM: Token-Level Quality for Self-Correction}
PRISM augments a pretrained MDM with an additional head $g_\phi$ that predicts token-level quality. For a partially generated sequence $y$, the target quantity is the probability that the current token $y_i$ is correct given its context:
\begin{equation}
    g_i^\star(y) := p(x_i = y_i \mid y \oplus m_i),
\end{equation}
where $y \oplus m_i$ denotes the sequence $y$ with the $i$-th token masked out. Intuitively, tokens that are inconsistent with their context should receive lower scores and become candidates for remasking.

The correction head is trained on partially denoised sequences, using a binary label that indicates whether a predicted token matches the ground truth. This objective is combined with the standard MDM loss, allowing the model to preserve its generation ability while learning a self-correction signal.

\subsection{PRISM Inference}

At inference time, PRISM maintains two sets of positions: masked positions, which are candidates for generation, and clean positions, which are candidates for remasking. Let $x^{(t)} = (x^{(t)}_1, \dots, x^{(t)}_L)$ denote the current partially denoised sequence at iteration $t$. 

The backbone diffusion model $f_\theta$ maps $x^{(t)}$ to a set of token-wise predictive distributions over the vocabulary:
\[
f_\theta(\cdot \mid x^{(t)})_i = p_\theta(x_i \mid x^{(t)}),
\]
while the auxiliary head $g_\phi$ produces a scalar quality score for each position $i$, denoted as $g_\phi(x^{(t)})_i$.

Thus, in a single forward pass, the model computes both token distributions and position-wise quality scores:
\begin{equation}
    f_\theta(\cdot \mid x^{(t)}),
    \qquad
    g_\phi(x^{(t)}).
\end{equation}

The decoder then selects a subset $T_t$ of low-quality clean positions to remask and a subset $S_t$ of masked positions to unmask, typically using a deterministic top-$K$ rule.

\begin{algorithm}[H]
\caption{PRISM inference step}
\label{alg:prism-infer}
\small
\begin{algorithmic}[1]
\State Given $x_{t_\ell}$, run one forward pass:
\Statex \hspace{\algorithmicindent} $f_\theta(\cdot \mid x_{t_\ell})$, $g_\phi(x_{t_\ell})$
\State Select $T_\ell$ (lowest-quality clean positions)
\State Select $S_\ell$ (masked positions to unmask)
\For{$j \in T_\ell$}
    \State $x_{t_\ell}^j \leftarrow m$
\EndFor
\For{$i \in S_\ell$}
    \State Sample $x_{t_{\ell+1}}^i \sim f_\theta^i(\cdot \mid x_{t_\ell})$
\EndFor
\end{algorithmic}
\end{algorithm}

\subsection{Why the Coupled Inference Rule Can Be Problematic}
The PRISM inference step is efficient, but it introduces two important limitations.

First, the procedure is \textbf{coupled}: remasking decisions and unmasking logits are computed from the same state $x_{t}$. As a result, tokens identified as low-quality still influence the distribution used to generate new tokens in the same step, which can lead to error propagation.

Second, the remasking rule is \textbf{greedy}. The model is observed to repetitively select the same lowest-scoring positions, making the correction process overly greedy. Importantly, low-quality scores may correspond not only to clearly incorrect tokens, but also to rare or context-specific tokens that are harder to predict. As a result, such tokens can be consistently removed even when they are valid, biasing the generation process. In long decoding runs, this behavior can reduce diversity and lead to mode collapse.

Our work is motivated by these two observations: the coupling between correction and generation, and the lack of stochasticity in the remasking policy.

\section{Methodology}

\subsection{Problem Formulation}
We study inference-time self-correction strategies for masked diffusion language models. Our goal is to improve the trade-off between generation quality and diversity without modifying either the pretrained diffusion backbone or the underlying training objective. This allows us to isolate the effect of the decoding policy itself.

Building on the formulation introduced in Section~3, we focus exclusively on modifications to the decoding procedure while keeping all training components fixed.

\subsection{Decoupled Remasking}

In the standard coupled decoding formulation, remasking and unmasking are both performed from the same intermediate state $x_t$. We instead consider a decoupled variant of the decoding step, where each iteration is divided into two stages:

\[
\begin{array}{l}
\text{Stage 1: } g_\phi(x^{(t)}) \Rightarrow T_t \subseteq C_t \\ 
\qquad\quad \tilde{x}^{(t)} \text{ obtained by masking } \{i \in T_t\} \\[4pt]
\text{Stage 2: } f_\theta(\cdot \mid \tilde{x}^{(t)}) \\ 
\qquad\quad x^{(t+1)}_i \sim f_\theta^i(\cdot \mid \tilde{x}^{(t)}),\; i \in S_t
\end{array}
\]

Here, $C_t$ denotes the set of clean (already revealed) positions. The subset $T_t \subseteq C_t$ is selected based on the quality scores $g_\phi(x_t)$. The intermediate state $\tilde{x}_t$ is obtained from $x_t$ by replacing all positions $i \in T_t$ with the mask token, while keeping the remaining positions unchanged.

This separation ensures that once a token is deemed unreliable and removed, it no longer influences the logits used to generate new tokens. Compared to the coupled procedure, this approach requires an additional forward pass but yields a cleaner self-correction mechanism.

\subsection{Stochastic Remasking}

Deterministic remasking follows a greedy top-$K_t$ policy, which often selects the same positions across iterations and can result in reduced diversity. To mitigate this, we introduce a stochastic remasking strategy inspired by temperature-based sampling in large language models. Instead of always choosing the lowest-scoring tokens, we sample positions according to a distribution induced by the quality scores and controlled by a temperature parameter. This allows us to balance focused correction and exploration during the decoding process.

We define a stochastic remasking policy over the set of clean tokens $C_t$:
\begin{equation}
\pi_i(\tau) = \frac{\exp(-s_i/\tau)}{\sum_{j \in C_t}\exp(-s_j/\tau)},
\end{equation}
where $s_i = g_\phi(x^{(t)})_i$ denotes the token-level quality logit produced by the correction head for position $i$ (i.e., an unnormalized score), and $\tau$ is a temperature parameter controlling the level of randomness. $K_t$ positions are selected without replacement according to the distribution $\pi(\tau)$.

\subsection{Temperature Scheduling}

A fixed temperature is generally suboptimal: early decoding benefits from conservative correction, while later stages can exploit additional stochasticity. Let
\begingroup
\scriptsize
\setlength{\abovedisplayskip}{2pt}
\setlength{\belowdisplayskip}{2pt}
\setlength{\jot}{1pt}
\begin{equation*}
\begin{gathered}
t\in\{0,\dots,T-1\},\qquad
u_t=\frac{t}{T-1}\in[0,1],\\
\Delta\tau=\tau_{\max}-\tau_{\min},\qquad
g_t=\tau_{t+1}-\tau_t,\quad t=0,\dots,T-2 .
\end{gathered}
\end{equation*}
\begin{equation*}
\begin{gathered}
0\le h,c,p_1,p_2<1,\qquad
p_1<p_2,\\
\tau_{\min}\le\tau_{\max},\qquad
\tau_{\min}\le\tau_{\rm mid}\le\tau_{\max}.
\end{gathered}
\end{equation*}
\endgroup

Here \(u_t\) denotes normalized decoding progress and \(g_t\) is the discrete scheduler increment.

\paragraph{Quadratic tail.}
\begingroup
\scriptsize
\setlength{\abovedisplayskip}{0pt}
\setlength{\belowdisplayskip}{1pt}
\renewcommand{\arraystretch}{0.86}
\begin{equation*}
\tau_t^{\rm quad}=
\begin{cases}
\tau_{\min},
& u_t\le h,\\

\tau_{\min}
+
\Delta\tau
\left(
\dfrac{u_t-h}{1-h}
\right)^2,
& u_t>h .
\end{cases}
\end{equation*}
\endgroup

\paragraph{Sigmoid.}
\begingroup
\scriptsize
\setlength{\abovedisplayskip}{0pt}
\setlength{\belowdisplayskip}{1pt}
\begin{equation*}
\tau_t^{\rm sig}
=
\tau_{\min}
+
\Delta\tau\,\sigma(k(u_t-c)),
\qquad
\sigma(z)=\frac{1}{1+\exp(-z)} .
\end{equation*}
\endgroup

\paragraph{Piecewise.}
\begingroup
\scriptsize
\setlength{\abovedisplayskip}{0pt}
\setlength{\belowdisplayskip}{1pt}
\renewcommand{\arraystretch}{0.86}
\begin{equation*}
\tau_t^{\rm pw}=
\begin{cases}
\tau_{\min},
& u_t\le p_1,\\

\tau_{\min}
+
(\tau_{\rm mid}-\tau_{\min})
\dfrac{u_t-p_1}{p_2-p_1},
& p_1<u_t\le p_2,\\

\tau_{\rm mid}
+
(\tau_{\max}-\tau_{\rm mid})
\left(
\dfrac{u_t-p_2}{1-p_2}
\right)^{1.5},
& u_t>p_2 .
\end{cases}
\end{equation*}
\endgroup

\begingroup
\emergencystretch=1em
All schedules are monotone non-decreasing in \(t\), implying \(g_t\ge0\). The quadratic and piecewise schedules concentrate temperature growth near the end of decoding, whereas the sigmoid schedule places the largest update around \(u_t=c\).
\par
\endgroup

\subsection{Proposed algorithm}

We summarize the full decoding procedure combining decoupled remasking, stochastic selection, and temperature scheduling.

\begin{algorithm}[H]
\caption{NAVIRA inference}
\label{alg:full}
\small
\begin{algorithmic}[1]
\State Initialize $x^{(0)} \leftarrow (m, \dots, m)$
\For{$t = 0, \dots, T-1$}
    \State Compute temperature $\tau_t$
    
    \State Compute scores: $s_i = g_\phi(x^{(t)})_i$
    \State $C_t \leftarrow \{i : x^{(t)}_i \neq m\}$ 
    
    \State $\pi_i \propto \exp(-s_i / \tau_t)$ for $i \in C_t$
    \State Sample $T_t \subseteq C_t$
    
    \State $\tilde{x}^{(t)}_i \leftarrow 
    \begin{cases}
    m, & i \in T_t \\
    x^{(t)}_i, & \text{otherwise}
    \end{cases}$
    \State $M_t \leftarrow \{i : \tilde{x}^{(t)}_i = m\}$
    \State Compute logits: $f_\theta(\cdot \mid \tilde{x}^{(t)})$
    \State Select $S_t \subseteq M_t$
    
    \For{$i \in S_t$}
        \State Sample $x^{(t+1)}_i \sim f_\theta^i(\cdot \mid \tilde{x}^{(t)})$
    \EndFor
\EndFor
\end{algorithmic}
\end{algorithm}

\section{Experiments}

We evaluate several inference-time remasking strategies for masked diffusion language models. Throughout this section, we use the following naming convention:

\begin{itemize}
    \item \textsc{PRISM}: the original deterministic PRISM remasking strategy;
    \item \textsc{PRISM} stochastic: a stochastic variant of the original PRISM inference procedure without decoupled forward passes. We include this variant to isolate the effect of stochastic token selection itself and verify that the observed improvements are not explained solely by introducing sampling randomness;
    \item \textsc{NAVIRA-det}: our deterministic decoupled remasking variant;
    \item \textsc{NAVIRA}: stochastic remasking with temperature-controlled token selection combined with decoupled inference;
    \item \textsc{NAVIRA} + scheduler: stochastic decoupled remasking with adaptive temperature scheduling.

\end{itemize}

We evaluate our inference strategies in the unconditional text generation setting using a pretrained 170M masked diffusion language model \citep{mdlm} trained on OpenWebText \citep{openwebtext}. Decoding starts from a fully masked sequence and proceeds iteratively for a fixed number of denoising steps. Unless stated otherwise, we evaluate generation across diffusion budgets ranging from 64 to 2048 steps and generate 1000 samples for each configuration.

Our experiments are designed as an ablation study rather than a benchmark comparison. The goal is to isolate the effect of inference-time remasking design while keeping the underlying model and training procedure fixed. We compare \textsc{PRISM}, \textsc{NAVIRA-det}, stochastic \textsc{NAVIRA} variants with fixed temperatures, adaptive scheduler-based \textsc{NAVIRA} variants, and decoding without remasking as a reference baseline. In addition, we compare against non-learned remasking strategies based on entropy and margin criteria inspired by Dream-style decoding \citep{dream7b}, where token selection is driven by uncertainty estimates rather than learned token-quality scores.

We report two complementary metrics. \textit{Perplexity} is used as a proxy for fluency and is computed with Qwen2.5-72B \citep{qwen25} as a reference language model: lower perplexity indicates that the generated text is more consistent with the distribution captured by a strong pretrained LLM. \textit{Entropy} is computed as the Shannon entropy of the empirical token distribution over generated samples and serves as a lightweight diversity diagnostic. In particular, entropy is useful for detecting mode collapse, where generations become overly concentrated or repetitive. We report both metrics because improvements in perplexity alone may come at the cost of reduced diversity.

\subsection{PRISM versus Non-learned Remasking Baselines}

We begin by comparing \textsc{PRISM} against greedy remasking strategies based on entropy and margin criteria, which do not use learned token-quality estimation. The goal of this comparison is to evaluate whether explicitly learned quality scores provide a more reliable correction signal than generic uncertainty-based heuristics.

Appendix~\ref{app:stochastic-no-temp} reveals a consistent pattern. All methods eventually exhibit signs of mode collapse as the number of forward passes increases, including both \textsc{PRISM} and \textsc{NAVIRA-det}. However, entropy- and margin-based remasking collapse substantially faster and produce less stable trade-offs between fluency and diversity as the decoding budget grows. While these heuristic strategies can improve perplexity, this improvement is accompanied by a rapid drop in entropy, indicating increasingly repetitive and concentrated generations.

In contrast, \textsc{PRISM} degrades more gradually, with \textsc{NAVIRA-det} consistently achieving the best overall balance between fluency and diversity among the compared methods. In particular, \textsc{NAVIRA-det} achieves lower perplexity than \textsc{PRISM} while maintaining substantially higher entropy than the uncertainty-based baselines. These results suggest that learned token-quality estimation provides a more reliable correction signal than generic uncertainty heuristics, and that the way this signal is incorporated into the decoding procedure is critical.

\subsection{Stochastic Remasking Without Temperature Scheduling}

We next examine whether entropy collapse can be mitigated by replacing deterministic top-$K$ remasking with stochastic sampling. Appendix~\ref{app:stochastic-no-temp} compares deterministic and stochastic variants of both \textsc{PRISM} and \textsc{NAVIRA-det}.

The results show that introducing stochasticity substantially increases entropy across all forward-pass budgets. In particular, fixed-temperature stochastic \textsc{NAVIRA} variants maintain higher diversity even in regimes where deterministic \textsc{NAVIRA-det} has already begun to collapse.

This improvement in diversity, however, comes at the cost of higher perplexity. These findings suggest that greedy remasking is a major contributor to collapse, but unconstrained stochasticity is not sufficient on its own: while it helps preserve diversity, it also reduces the fluency gains that remasking is intended to provide.

\begin{figure*}[h]
    \centering
    \includegraphics[width=0.90\textwidth]{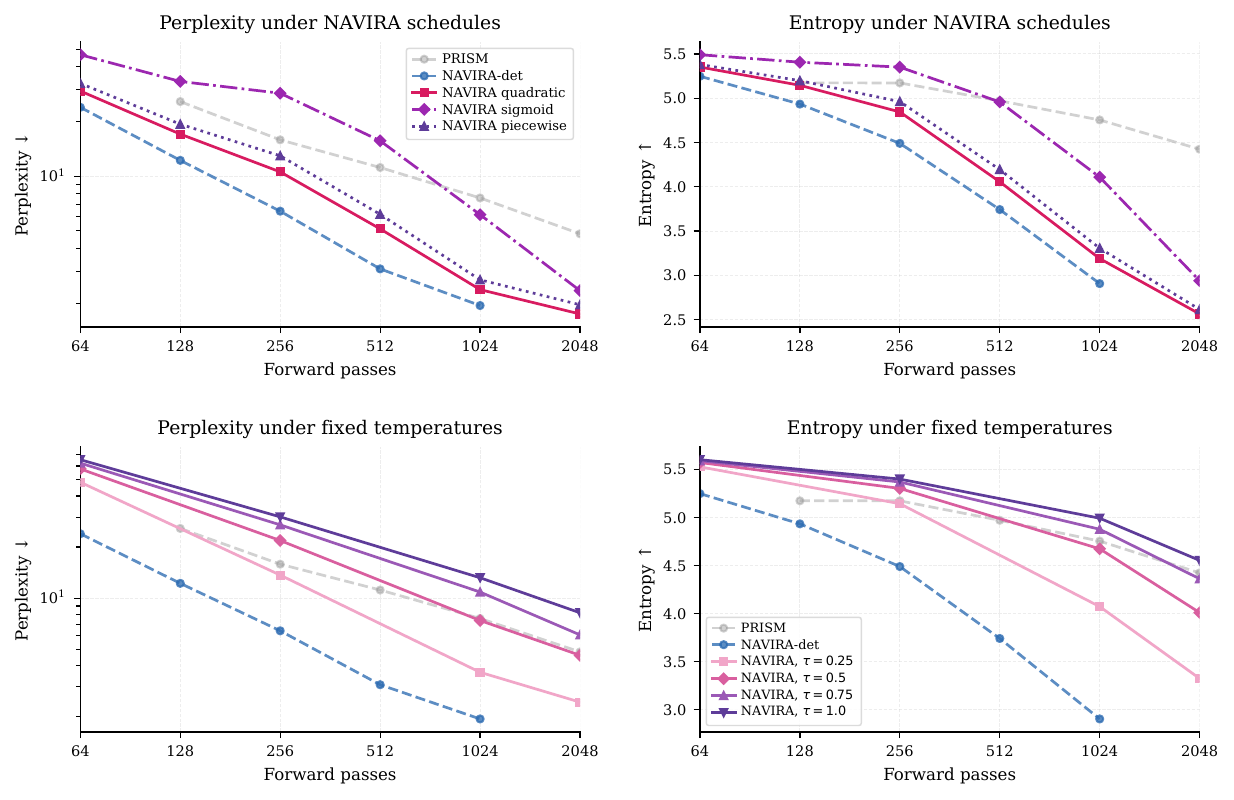}
    \caption{Temperature-controlled stochastic remasking. \textsc{NAVIRA-det} corresponds to deterministic remasking, while \textsc{NAVIRA} uses stochastic token selection with either fixed temperatures or adaptive schedulers. Lower temperatures behave more greedily and preserve better perplexity, whereas higher temperatures maintain greater diversity. Adaptive schedulers further improve this balance by gradually increasing stochasticity throughout decoding.}
    \label{fig:temperature-sweep}
\end{figure*}

\subsection{Temperature-Controlled Stochastic Remasking}

The preceding experiment shows that stochastic remasking mitigates entropy collapse, but excessive stochasticity can substantially degrade perplexity. We therefore study how the strength of stochastic remasking should be controlled throughout the decoding trajectory. Figure ~\ref{fig:temperature-sweep} jointly compares deterministic remasking in \textsc{NAVIRA-det}, fixed-temperature stochastic remasking in \textsc{NAVIRA}, and several adaptive temperature schedulers.

The overall trend is consistent across inference budgets. Lower temperatures concentrate sampling on low-quality tokens, producing behavior closer to deterministic remasking in \textsc{NAVIRA-det} and yielding lower perplexity. Higher temperatures flatten the sampling distribution, increasing exploration and preserving higher entropy. As a result, no single fixed temperature is uniformly optimal: the preferred operating point depends on the available forward-pass budget and on the desired balance between fluency and diversity.

Motivated by this observation, we further investigate adaptive temperature schedules that vary the degree of stochasticity during decoding. Early decoding stages benefit from lower temperatures, which encourage decisive correction while the sequence remains unstable, whereas larger temperatures become more useful later in the trajectory to avoid convergence to narrow modes. Although the quantitative differences between scheduler families remain smaller than the gap between deterministic and stochastic remasking overall, gradually increasing schedules consistently provide the most favorable quality--diversity trade-off.

To summarize the trends observed across diverse inference budgets, Table~\ref{tab:main-results-owt} reports the main quantitative results for a representative subset of decoding strategies, highlighting their respective quality--diversity trade-offs.

\begin{table*}[h]
\centering
\small
\begin{tabular}{lcc cc cc cc cc}
\toprule
& \multicolumn{2}{c}{64 forwards}
& \multicolumn{2}{c}{128 forwards}
& \multicolumn{2}{c}{256 forwards}
& \multicolumn{2}{c}{512 forwards}
& \multicolumn{2}{c}{1024 forwards} \\
\cmidrule(lr){2-3}\cmidrule(lr){4-5}\cmidrule(lr){6-7}
\cmidrule(lr){8-9}\cmidrule(lr){10-11}
Method
& PPL $\downarrow$ & Ent. $\uparrow$
& PPL $\downarrow$ & Ent. $\uparrow$
& PPL $\downarrow$ & Ent. $\uparrow$
& PPL $\downarrow$ & Ent. $\uparrow$
& PPL $\downarrow$ & Ent. $\uparrow$ \\
\midrule

ReMDM
& 72.51 & 3.72
& 16.69 & 1.80
& 9.71 & 1.10
& 16.74 & 1.04
& 24.44 & 1.68 \\

MDLM backbone
& 89.57 & 5.69
& 76.30 & 5.66
& 72.17 & 5.65
& 69.97 & 5.62
& 67.01 & 5.62 \\

\midrule

\textsc{PRISM}
& -- & --
& 25.75 & 5.17
& 15.85 & 5.17
& 11.17 & 4.97
& 7.59 & 4.75 \\

\textsc{NAVIRA-det}
& 23.95 & 5.25
& 12.22 & 4.93
& 6.42 & 4.49
& 3.09 & 3.74
& 1.94 & 2.91 \\

\textbf{\textsc{NAVIRA} + piecewise}
& 32.43 & 5.38
& 19.34 & 5.20
& 12.94 & 4.96
& 6.17 & 4.19
& 2.69 & 3.30 \\

\bottomrule
\end{tabular}

\caption{
Main results on OpenWebText. We report generation perplexity and entropy for different inference forward budgets. Lower perplexity indicates higher fluency, while higher entropy reflects better diversity. Boldface highlights the configuration achieving the strongest overall quality--diversity trade-off among the proposed methods across both metrics.
}
\label{tab:main-results-owt}
\end{table*}

\subsection{Correction Trajectory Analysis}

The previous experiments demonstrate that different remasking strategies lead to distinct quality--diversity trade-offs. We next investigate whether these differences are reflected directly in the decoding dynamics themselves. Rather than comparing only final generations, we analyze the correction trajectories induced by different remasking policies, i.e., the sets of token positions selected for remasking at each denoising step. 

For each decoding step $t$, we record the remasked token set $T_t$ and compare trajectories across methods using set-overlap statistics, including intersection size, Jaccard similarity, and overlap coefficient.
Table~\ref{tab:trajectory-overlap} summarizes the pairwise overlap statistics. \textsc{PRISM} and \textsc{NAVIRA-det} exhibit relatively small overlap, indicating that decoupled remasking already induces noticeably different correction dynamics. \textsc{NAVIRA} diverges substantially further: both comparisons involving stochastic remasking yield near-zero Jaccard similarity, showing that stochastic remasking selects almost entirely different token subsets throughout decoding.

Most importantly, the comparison between \textsc{NAVIRA-det} and stochastic \textsc{NAVIRA} demonstrates that stochastic remasking is not merely a minor perturbation of deterministic decoding. Instead, it produces fundamentally different correction trajectories, confirming that the remasking policy changes not only the final generations but also the iterative refinement process itself.

\begin{table}[t]
\centering
\scriptsize
\setlength{\tabcolsep}{3pt}
\begin{tabular}{llccc}
\toprule
Method A & Method B & Inter. & Jaccard & Overlap \\
\midrule

PRISM & NAVIRA-det
& 0.716 & 5.91\% & 19.23\% \\

PRISM & NAVIRA ($\tau=0.2$)
& 0.062 & 0.33\% & 4.67\% \\

NAVIRA-det & NAVIRA ($\tau=0.2$)
& 0.116 & 0.48\% & 3.61\% \\

\bottomrule
\end{tabular}

\caption{
Pairwise overlap between remasking trajectories. Stochastic NAVIRA produces nearly non-overlapping correction trajectories compared to deterministic variants.
}
\label{tab:trajectory-overlap}
\end{table}

\subsection{LLM-based Generation Evaluation}

While perplexity and entropy provide useful automatic diagnostics, they do not fully capture qualitative properties of generated text. We therefore additionally evaluate generations using an LLM-as-a-judge protocol \citep{zheng2023judgingllmasajudgemtbenchchatbot}. As a reference evaluator, we use Qwen3-235B-A22B-Instruct-2507, which scores generated samples along several dimensions, including clarity, grammaticality, factuality, style, and creativity.  The full evaluation
prompt is provided in Appendix~\ref{app:llm-judge-prompt}.

For each configuration, we generate 1000 samples and ask the judge model to assign scalar quality scores using a fixed evaluation prompt. Appendix~\ref{app:llm-judge} summarizes the resulting scores across different forward pass budgets.

Among the evaluated configurations, scheduler-based stochastic \textsc{NAVIRA} achieves the strongest overall subjective generation quality. In particular, \textsc{NAVIRA} with sigmoid temperature scheduling achieves the highest overall score across nearly all decoding budgets, while also obtaining the strongest grammaticality and creativity ratings. Interestingly, this trend differs from the perplexity-based evaluation, where deterministic \textsc{NAVIRA-det} often achieved the best fluency scores. A possible explanation is that stochastic remasking introduces more globally coherent and stylistically diverse revisions, which are preferred by instruction-tuned LLM judges even when token-level likelihood slightly deteriorates.

Overall, these results suggest that stochastic remasking improves not only diversity metrics, but also the perceived subjective quality of generations under strong LLM-based evaluation.

\section{Conclusion}
This work demonstrates that the performance of PRISM-style self-correction in masked diffusion language models is governed not only by the availability of learned token-level quality estimates, but also by the decoding policy through which those estimates are used. Our controlled ablations show that decoupling remasking from token generation yields more reliable correction by preventing tokens identified as unreliable from conditioning subsequent predictions. Deterministic remasking is further shown to be prone to inducing repetitive correction behavior and entropy collapse, whereas stochastic remasking improves diversity by reducing repeated selection of the same low-scoring positions. Finally, temperature control, particularly through scheduled stochasticity, provides a principled mechanism for balancing fluency and diversity across different inference budgets. Overall, these results highlight inference-time remasking design as a significant factor in the reliability and flexibility of diffusion-based text generation, and demonstrate that the proposed algorithm can substantially enhance the quality-diversity trade-off without modifying the underlying model or training objective.

\section*{Limitations}
Our evaluation is limited to a 170M masked diffusion language model in the unconditional OpenWebText generation setting, so the observed trends may not directly transfer to larger models or stronger diffusion backbones. In addition, we evaluate the proposed remasking strategies only in unconditional text generation and do not study their behavior on downstream conditional tasks such as summarization, code generation, or instruction following, where the quality-diversity trade-off and correction dynamics may differ substantially.


\bibliography{custom}

\clearpage

\appendix
\onecolumn

\setcounter{section}{0}
\renewcommand{\thesection}{\Alph{section}}
\renewcommand{\thesubsection}{\Alph{section}.\arabic{subsection}}

\newcommand{\appsection}[2]{%
  \refstepcounter{section}%
  \setcounter{subsection}{0}%
  \section*{Appendix \thesection. #1}%
  \label{#2}%
}

\appsection{Additional Remasking Comparisons}{app:stochastic-no-temp}

\begin{figure*}[h]
    \centering
    \includegraphics[width=0.90\textwidth]{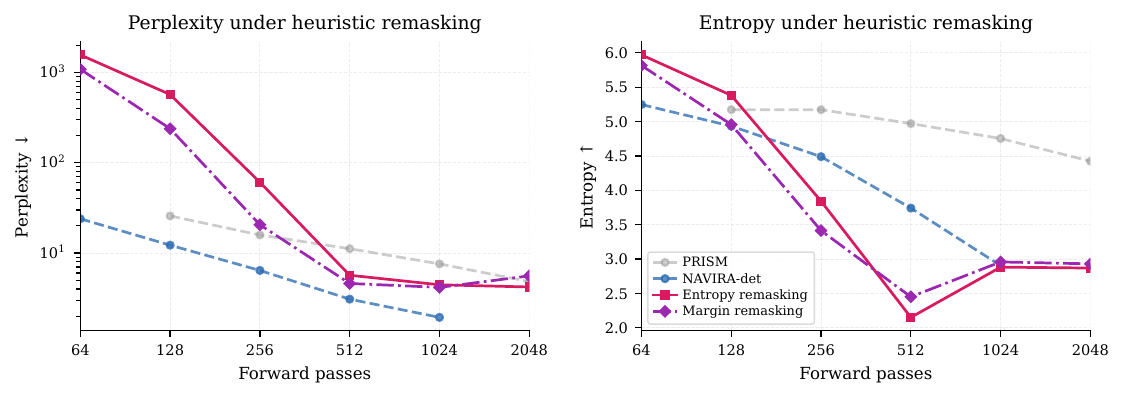}
    \caption{Comparison between \textsc{PRISM}, \textsc{NAVIRA-det}, and basic Dream-style remasking baselines on the 170M MDM. \textsc{NAVIRA-det} achieves the best overall quality--diversity trade-off, while the entropy- and margin-based heuristics exhibit strong entropy collapse as the number of forward passes grows.}
\end{figure*}

\begin{figure*}[h]
    \centering
    \includegraphics[width=0.90\textwidth]{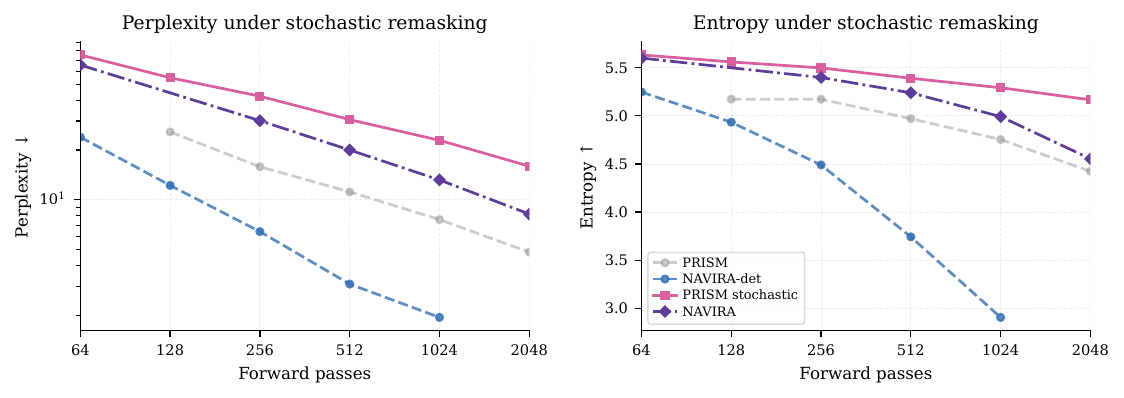}
    \caption{Deterministic versus stochastic remasking without temperature scheduling. Stochastic \textsc{NAVIRA} substantially increases entropy for both \textsc{PRISM} and \textsc{NAVIRA-det}, but incurs a noticeable perplexity cost, indicating that naive stochasticity alone does not resolve the quality--diversity trade-off.}
\end{figure*}

\FloatBarrier

\appsection{LLM-as-a-Judge Evaluation Prompt}{app:llm-judge-prompt}

For LLM-based evaluation, we use the following fixed prompt for all generated samples:

\begin{verbatim}
You are an impartial expert evaluator for generated text.
Evaluate the following text on a scale from 1 to 10 for:
1. Clarity
2. Grammaticality
3. Factuality
4. Style
5. Creativity

Generated text:
{text}

Return ONLY valid JSON in exactly this format:
{
  "clarity": float,
  "grammaticality": float,
  "factuality": float,
  "style": float,
  "creativity": float
}
\end{verbatim}

\FloatBarrier

\appsection{Additional LLM-as-a-Judge Results}{app:llm-judge}

\begin{figure*}[h]
    \centering
    \includegraphics[width=0.95\textwidth]{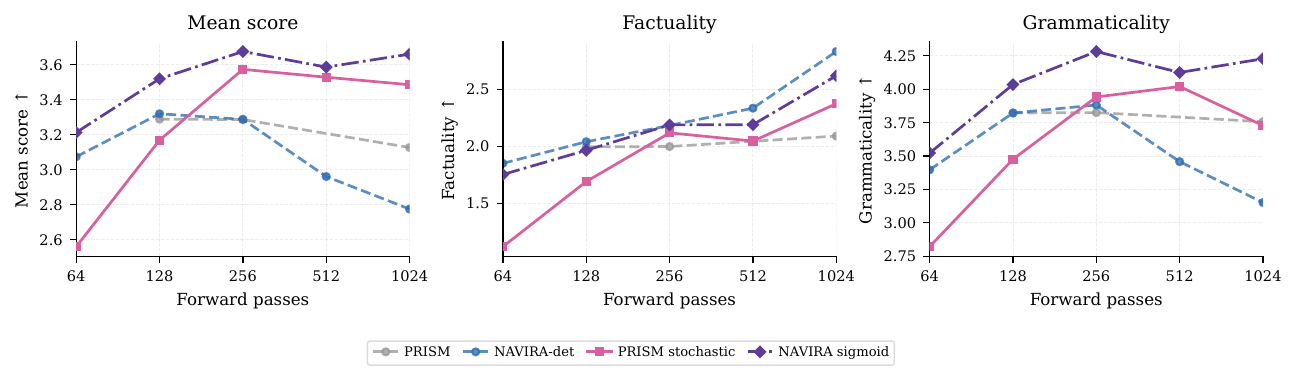}
    \caption{LLM-as-a-judge evaluation using Qwen3-235B-A22B-Instruct-2507. Stochastic \textsc{NAVIRA} with sigmoid scheduling achieves the strongest overall scores across most inference budgets, particularly in grammaticality and overall generation quality.}
    \label{fig:llm-judge}
\end{figure*}

\begin{table*}[!ht]
\centering
\scriptsize
\setlength{\tabcolsep}{4pt}
\renewcommand{\arraystretch}{1}
\begin{tabular}{l|c|ccccc}
Method
& Mean score
& Clarity
& Grammar
& Factuality
& Style
& Creativity \\
\midrule

\multicolumn{7}{c}{64 forward passes} \\
\midrule

\textsc{NAVIRA} + sigmoid
& \textbf{3.212}
& \textbf{2.630}
& \textbf{3.522}
& 1.752
& \textbf{3.465}
& \textbf{4.692} \\

\textsc{NAVIRA-det}
& 3.073
& 2.605
& 3.397
& \textbf{1.850}
& 3.180
& 4.332 \\

\textsc{PRISM} stochastic
& 2.559
& 2.080
& 2.822
& 1.118
& 2.748
& 4.026 \\

\midrule
\multicolumn{7}{c}{128 forward passes} \\
\midrule

\textsc{NAVIRA} + sigmoid
& \textbf{3.519}
& \textbf{2.980}
& \textbf{4.035}
& 1.965
& \textbf{3.780}
& \textbf{4.835} \\

\textsc{NAVIRA-det}
& 3.319
& 2.850
& 3.822
& \textbf{2.040}
& 3.417
& 4.465 \\

\textsc{PRISM}
& 3.288
& 2.690
& 3.822
& 1.995
& 3.488
& 4.442 \\

\textsc{PRISM} stochastic
& 3.168
& 2.560
& 3.476
& 1.690
& 3.424
& 4.688 \\

\midrule
\multicolumn{7}{c}{256 forward passes} \\
\midrule

\textsc{NAVIRA} + sigmoid
& \textbf{3.675}
& \textbf{3.120}
& \textbf{4.282}
& \textbf{2.188}
& 3.905
& 4.883 \\

\textsc{PRISM} stochastic
& 3.573
& 2.880
& 3.942
& 2.118
& \textbf{3.910}
& \textbf{5.016} \\

\textsc{NAVIRA-det}
& 3.287
& 2.795
& 3.882
& 2.183
& 3.283
& 4.290 \\

\textsc{PRISM}
& 3.286
& 2.700
& 3.825
& 1.998
& 3.493
& 4.415 \\

\midrule
\multicolumn{7}{c}{512 forward passes} \\
\midrule

\textsc{NAVIRA} + sigmoid
& \textbf{3.586}
& \textbf{2.990}
& \textbf{4.125}
& 2.190
& \textbf{3.808}
& 4.817 \\

\textsc{PRISM} stochastic
& 3.528
& 2.932
& 4.020
& 2.046
& 3.706
& \textbf{4.936} \\

\textsc{NAVIRA-det}
& 2.960
& 2.355
& 3.458
& \textbf{2.335}
& 2.688
& 3.965 \\

\midrule
\multicolumn{7}{c}{1024 forward passes} \\
\midrule

\textsc{NAVIRA} + sigmoid
& \textbf{3.660}
& \textbf{3.000}
& \textbf{4.228}
& 2.620
& \textbf{3.752}
& 4.702 \\

\textsc{PRISM} stochastic
& 3.485
& 2.804
& 3.728
& 2.374
& 3.350
& \textbf{5.168} \\

\textsc{PRISM}
& 3.126
& 2.510
& 3.757
& 2.090
& 3.058
& 4.217 \\

\textsc{NAVIRA-det}
& 2.774
& 2.125
& 3.152
& \textbf{2.833}
& 2.303
& 3.458 \\

\bottomrule
\end{tabular}

\caption{
LLM-as-a-judge evaluation using Qwen3-235B-A22B-Instruct-2507. We report average subjective scores over 1000 generated samples for each decoding configuration. Stochastic \textsc{NAVIRA} with sigmoid scheduling consistently achieves the strongest overall quality assessments across most inference budgets.
}

\label{tab:llm-judge}
\end{table*}

\end{document}